\documentclass[letterpaper]{article} 
\usepackage{format} 

\usepackage{times}  
\usepackage{helvet}  
\usepackage{courier}  
\usepackage[hyphens]{url}  
\usepackage{graphicx} 
\usepackage{natbib}  
\usepackage{caption} 
\usepackage{algorithm}
\usepackage{algorithmic}
\usepackage{amsmath}
\usepackage{booktabs}

\usepackage{bm}
\usepackage{newfloat}
\usepackage{listings}
\DeclareCaptionStyle{ruled}{labelfont=normalfont,labelsep=colon,strut=off} 
\floatstyle{ruled}
\newfloat{listing}{tb}{lst}{}
\floatname{listing}{Listing}

\title{Quantum-Enhanced Multi-Task Learning with Learnable Weighting \\for Pharmacokinetic and Toxicity Prediction}

\author{
  Han Zhang, Fengji Ma, Jiamin Su, \\
  Xinyue Yang, Lei Wang, Wen-Cai Ye, Li Liu
}
\date{}
\begin{document}

\maketitle

\begin{abstract}

Prediction for ADMET (Absorption, Distribution, Metabolism, Excretion, and Toxicity) plays a crucial role in drug discovery and development, accelerating the screening and optimization of new drugs. Existing methods primarily rely on single-task learning (STL), which often fails to fully exploit the complementarities between tasks. Besides, it requires more computational resources while training and inference of each task independently. To address these issues, we propose a new unified \textbf{Q}uantum-enhanced and task-\textbf{W}eighted  \textbf{M}ulti-\textbf{T}ask  \textbf{L}earning (\textbf{QW-MTL}) framework, specifically designed for ADMET classification tasks. Built upon the Chemprop-RDKit backbone, QW-MTL adopts quantum chemical descriptors to enrich molecular representations with additional information about the electronic structure and interactions. Meanwhile, it introduces a novel exponential task weighting scheme that combines dataset-scale priors with learnable parameters to achieve dynamic loss balancing across tasks. To the best of our knowledge, this is the first work to systematically conduct joint multi-task training across all 13 Therapeutics Data Commons (TDC) classification benchmarks, using leaderboard-style data splits to ensure a standardized and realistic evaluation setting. Extensive experimental results show that QW-MTL significantly outperforms single-task baselines on 12 out of 13 tasks, achieving high predictive performance with minimal model complexity and fast inference, demonstrating the effectiveness and efficiency of multi-task molecular learning enhanced by quantum-informed features and adaptive task weighting.
\end{abstract}
\section{Introduction}

Predicting ADMET (Absorption, Distribution, Metabolism, Excretion, and Toxicity) properties is essential for early-stage drug discovery, enabling the identification of candidate molecules with favorable pharmacokinetic and safety profiles~\cite{1,2,4}. Given the high cost and time required for laboratory-based ADMET testing, machine learning (ML) models have been essential tools for computational ADMET prediction~\cite{4,5}.

Despite a growing body of work, traditional methods based on single-task learning remain highly competitive in ADMET prediction. For example, the fingerprint-based gradient boosting frameworks by Tian et al.~\cite{44} and Notwell et al.~\cite{45} achieve top performance across multiple ADMET tasks from Therapeutics Data Commons (TDC) benchmark~\cite{3}, which is a widely used and standardized platform for machine learning in drug discovery that provides curated datasets and standardized evaluation protocols. Although single-task models can achieve strong performance in certain scenarios, their isolated training paradigm limits their ability to generalize across diverse molecular properties and often incurs higher inference time.

Currently, multi-task learning (MTL) has emerged as a promising strategy to improve data generalization and efficiency by leveraging shared information across tasks and jointly predicting multiple molecular properties~\cite{6}. However, its application to ADMET prediction remains limited and its benefits inconclusive~\cite{7}. To summarize, there are two key challenges hinder its broader adoption: \textbf{Firstly}, different ADMET tasks may rely on distinct structural or physicochemical aspects of a molecule, making it difficult to construct a unified and sufficiently expressive representation that supports effective knowledge sharing across tasks. \textbf{Secondly}, due to large heterogeneity in task objectives, data sizes, and learning difficulties, imbalanced optimization and inter-task interference often arise during training, potentially degrading overall performance. Moreover, most prior multi-task learning studies either evaluate on a small subset of tasks or rely on cross-validation and custom internal splits~\cite{8,9}, lacking standardized training and testing protocols. These evaluation strategies make it difficult to fairly assess generalization and may lead to inflated performance estimates.

To address these challenges and establish a standardized benchmark for ADMET multi-task evaluation, we present a novel \textbf{Q}uantum-enhanced and task-\textbf{W}eighted  \textbf{M}ulti-\textbf{T}ask  \textbf{L}earning framework (\textbf{QW-MTL}). It is a systematic investigation of MTL on all 13 ADMET classification tasks from the TDC benchmark, strictly adhering to the official leaderboard train-test splits for unified training and evaluation. To the best of our knowledge, this is the first study to train and evaluate a unified multi-task model under this rigorous and realistic setting, providing a reliable baseline and evaluation paradigm for the field.

Building on this standardized framework, we tackle two key technical challenges in MTL. First, to overcome the limitations of conventional 2D molecular descriptors (e.g., those computed by RDKit), we introduce quantum chemical (QC) descriptors. These physically-grounded 3D features capture molecular spatial conformation and electronic properties that are essential for ADMET outcomes \cite{10,11}, thereby providing a richer, physically-informed representation. Second, to mitigate task interference (a common issue where heterogeneous task difficulties and data scales hinder joint optimization), we propose an adaptive task weighting mechanism, which dynamically adjusts each task's contribution to the total loss via a learnable, softplus-transformed $\beta$ vector. This allows the model to intelligently balance competing objectives during optimization, leading to improved stability and overall performance.

Our main contributions can be summarized as follows:
\begin{itemize}
\item \textbf{A Unified QW-MTL Framework.} We propose QW-MTL, a unified multi-task learning framework specifically designed for ADMET classification. To ensure fairness and reproducibility, we conduct the first systematic study across all 13 TDC ADMET classification tasks using the official leaderboard-style splits for joint training and evaluation.
\item \textbf{Quantum-Informed Representations and Adaptive Task Weighting.} To tackle the representation and optimization challenges in MTL, QW-MTL incorporates quantum chemical descriptors to enrich input features, and introduces a novel exponential task weighting mechanism that combines dataset-scale priors with learnable parameters for dynamic loss balancing.

\item \textbf{State-of-the-Art (SOTA) Performance.} We demonstrate through extensive experiments that QW-MTL significantly outperforms strong single-task baselines on 12 out of 13 tasks, validating the effectiveness of our approach and setting a new SOTA for multi-task ADMET prediction on this benchmark.
\end{itemize}

\section{Related Works}

\subsection{MTL for Molecular Property Prediction}
Relevant methods for ADMET prediction have evolved from traditional quantitative structure-activity relationship (QSAR) models to sophisticated deep learning architectures~\cite{18,19}. In particular, Graph Neural Networks (GNNs), such as Message Passing Neural Networks (MPNNs), have become the state-of-the-art for their ability to learn complex patterns from molecular graphs~\cite{11,28,29}. However, a prevalent issue is that these advanced models are almost universally applied under a single-task learning paradigm, which isolates the inherent connections between tasks, failing to leverage shared biochemical information. These limitations motivate the use of MTL, which aims to improve predictive performance and generalization by sharing representations across related tasks and adaptively balancing task contributions~\cite{6,30}. Pioneering works, including GROVER, have demonstrated on general benchmarks like MoleculeNet that MTL can significantly benefit low-resource tasks~\cite{31,32}. Some studies have attempted to apply MTL to ADMET prediction tasks~\cite{48}, particularly by sharing knowledge across different ADMET endpoints to improve model generalization. For example, MTGL-ADMET attempts to explore optimal task combinations for predicting the main task~\cite{49}. However, the efficacy of MTL in ADMET prediction remains inconclusive and faces several challenges. First, different ADMET tasks vary in complexity, data availability, and prediction targets, making it challenging to effectively leverage inter-task knowledge sharing. Second, Many ADMET tasks exhibit significant differences in data scales, which leads to inconsistent transfer learning effects across tasks. Third, the lack of standardized evaluation protocols leads to inconsistencies in performance comparison across models, often resulting in over-optimistic estimates of generalization. 

Recently, the directed message passing neural network (D-MPNN) adopted by Chemprop has demonstrated strong performance in molecular property prediction~\cite{11,43}. Furthermore, ADMET-AI enhances predictive performance by integrating Chemprop with molecular descriptors computed using RDKit~\cite{9}. In this work, we adopt the combined model (Chemprop + RDKit) as our backbone and use its single-task learning performance as a strong baseline, upon which we further explore enhanced strategies for multi-task joint training and optimization to better exploit inter-task synergies and improve overall predictive performance.

\begin{figure*}[!h]
  \centering    
  \includegraphics[width=1\textwidth]{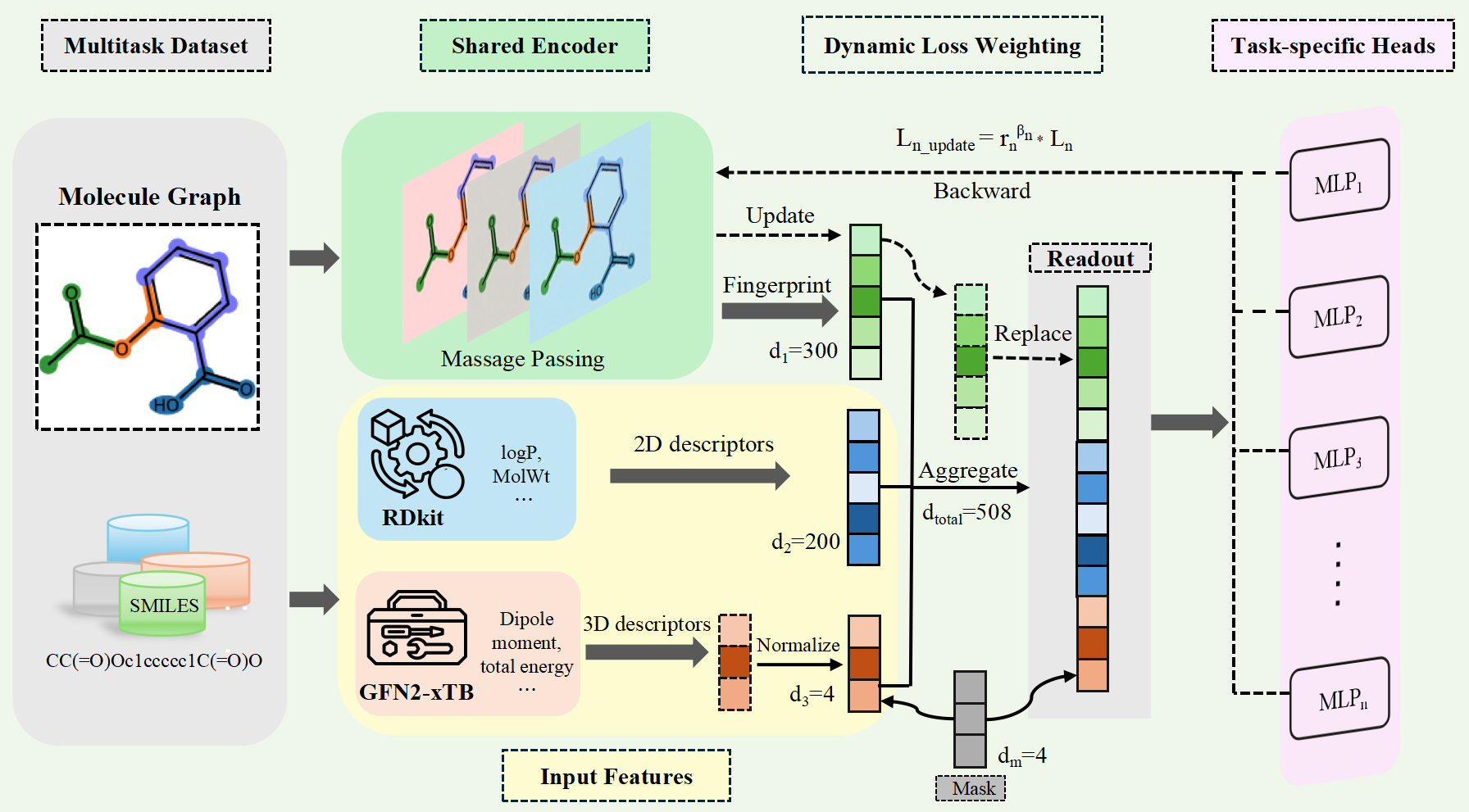}
  \caption{The framework of QW-MTL: All tasks share a common encoder (D-MPNN), and the learned molecular fingerprints are concatenated with additional input features. These combined features are then passed into each task's respective FFN layer. The loss is backpropagated through learnable task weights, and the final prediction of ADMET properties is made.}
  \label{fig:framework}
  \vspace{-4mm}
\end{figure*}

\subsection{Quantum-informed Molecular Representation}
The quality of molecular representation is paramount for predictive performance. Traditional 2D representations, such as graphs and fingerprints, while efficient, neglect 3D conformational and electronic properties that are crucial for intermolecular interactions \cite{33,34,35}. These physicochemical properties are especially vital for accurately predicting ADMET endpoints like solubility and permeability.
To bridge this gap, recent studies have started incorporating quantum chemical descriptors to provide physically-grounded insights into molecular behavior~\cite{28,36}. Although previous studies~\cite{10} have incorporated quantum chemical descriptors into ADMET prediction, employing the graph transformer architecture, our work broadens the scope of quantum descriptors by introducing four types of quantum features---dipole moment, HOMO-LUMO gap, electrons, and total energy and adopts the powerful D-MPNN model architecture, which shows strong potential for the model's predictive performance and generalization ability.
\subsection{Task Weighting in MTL}

A central challenge in MTL is how to balance the relative importance of each task during training. Simply averaging task losses can lead to imbalanced learning, where dominant tasks suppress weaker ones and degrade overall performance \cite{46}. This issue is especially pronounced in molecular modeling due to large variations in task scale, difficulty, and label sparsity.

Various task-weighting strategies have been proposed to address this. Uncertainty-based methods~\cite{40} assign weights based on task confidence. Gradient-balancing techniques such as GradNorm~\cite{41} dynamically adjust task weights to equalize gradient magnitudes. TP-AMTL~\cite{42} explores the concept of negative transfer in MTL and emphasize that task weights can dynamically adjust to mitigate the adverse effects of task interactions. While these methods have shown promise in some domains, they often require auxiliary objectives or careful tuning, and may not generalize well to ADMET classification tasks, where task interference is minimal and heterogeneity is more strongly driven by data scale. To address these limitations, we propose the first method that applies a weighting strategy based on sample scale to ADMET multi-task learning, using a simple, learnable exponent for weighting. This makes our framework more flexible, efficient, and better aligned with the practical demands of multi-task molecular modeling.
\section{Methodology}
\subsection{Overview of QW-MTL Framework}
Figure~\ref{fig:framework} presents the overall architecture of the proposed QW-MTL framework. Given SMILES strings from a multi-task ADMET dataset, we first employ a shared D-MPNN encoder to learn molecular fingerprints that capture the 2D structural information. In parallel, we extract expert-crafted 2D physicochemical descriptors (e.g., logP, MolWt) using RDKit~\cite{55}, and compute 3D quantum chemical descriptors (e.g., HOMO–LUMO gap, dipole moment, total energy) using GFN2-xTB ~\cite{56}. These three types of features are concatenated to form a unified molecular representation. This fused representation is then fed into task-specific feedforward networks (FFNs), each responsible for predicting a particular ADMET property. During training, we introduce a data-scale-aware task weighting mechanism, where each task's loss is scaled by a learnable weight based on its sample size. The model is optimized via backpropagation using the aggregated weighted loss. After training, the framework outputs ADMET property predictions for molecules in the test set.
\subsection{D-MPNN Architecture and RDKit Descriptors}
We adopt the architecture of ADMET-AI~\cite{9}, a recent SOTA on the TDC ADMET leaderboard, as our foundation. ADMET-AI builds upon Chemprop by integrating two main components: a directed message passing neural network (D-MPNN) and handcrafted physicochemical descriptors from RDKit. This hybrid structure has been shown to substantially improve predictive accuracy over using D-MPNN alone, particularly on endpoints such as solubility and permeability that are sensitive to global molecular properties.

We follow this design and use Chemprop’s D-MPNN as the shared encoder to learn structure-based molecular fingerprints from SMILES inputs. D-MPNN operates on directed molecular graphs, where messages are passed along bonds rather than atoms, reducing over-smoothing and capturing direction-aware chemical interactions. In parallel, we compute a set of RDKit-based 2D descriptors (e.g., logP, molecular weight, number of rotatable bonds), which provide global physicochemical properties that are complementary to local structural features captured by D-MPNN. These descriptors are concatenated with the learned fingerprint $\mathbf{z}$ to form an enriched molecular representation, which is then used in downstream multi-task prediction.

\subsection{Quantum Feature Integration}

While 2D structural features and physicochemical descriptors provide valuable information for ADMET prediction, they often neglect quantum-level electronic properties that are critical for modeling molecular interactions, reactivity, and permeability. To further enrich molecular representation with physically grounded information, we incorporate quantum chemical descriptors computed using the GFN2-xTB toolkit~\cite{50}. GFN2-xTB is a semi-empirical tight-binding quantum chemical method that achieves a good balance between computational efficiency and physical accuracy. Compared to full ab initio~\cite{51} or density functional theory (DFT)~\cite{52} methods, it offers a much lower computational cost while maintaining acceptable precision, making it particularly suitable for large-scale molecular datasets.

Specifically, we extract four quantum descriptors for each molecule: (1) dipole moment norm, (2) HOMO–LUMO gap, (3) total number of electrons, and (4) total electronic energy. These quantum-level features complement the local topological information learned by the D-MPNN encoder and the global physicochemical properties captured by RDKit, enhancing the model's ability to predict ADMET-related properties such as permeability and reactivity.

It is worth noting that not all SMILES strings yield valid quantum descriptors, due to occasional failures in 3D conformer optimization or quantum convergence. Across the entire dataset, the average extraction success rate is approximately 90\%. To ensure consistent input dimensions and preserve all samples aligned with the TDC task splits, we do not discard molecules with missing quantum features. Instead, we introduce a 4-dimensional binary mask to indicate which quantum values are missing, enabling the model to maintain training consistency and robustness in the presence of incomplete data.

Finally, the quantum descriptor vector and its corresponding mask are concatenated with the D-MPNN fingerprint and the 200-dimensional RDKit feature vector to form an enriched molecular representation. This fused input is then passed into task-specific predictors for downstream multi-task ADMET property prediction.

\subsection{Task Weighting Strategy Based on Sample Scale}
To address the problem of task imbalance in multi-task ADMET prediction, we propose a dynamic task weighting strategy based on sample scale. This method assigns a learnable exponent to each task, allowing the model to adjust the contribution of each task’s loss according to its sample proportion within each training batch.

Let $\mathcal{T} = \{1, 2, \dots, T\}$ denote the set of tasks. In each training batch, we count the number of valid (non-missing) labels $n_t$ for task $t \in \mathcal{T}$ and compute its sample proportion:
\begin{equation}
r_t = \frac{n_t}{\sum_{i=1}^{T} n_{i}} ,
\label{eq:rt}
\end{equation}

We then introduce a learnable log-exponent parameter $\log \beta_t$ for each task, and use the softplus activation to ensure a positive exponent. The task weight is defined as:
\begin{equation}
w_t = r_t^{\,\text{softplus}(\log \beta_t)} ,
\label{eq:wt}
\end{equation}

Given the mean binary cross-entropy loss $\mathcal{L}_t$ for each task, the total multi-task loss is computed as:
\begin{equation}
\mathcal{L}_{\text{total}} = \sum_{t=1}^{T} w_t \cdot \mathcal{L}_t ,
\label{eq:mtl_loss}
\end{equation}

In contrast to uniform weighting or uncertainty-based weighting methods, our approach is entirely data-driven and does not rely on manually specified task priors. By learning the exponent $\beta_t$ jointly with model parameters, the model can dynamically adapt to batch-level variations in label availability. This makes our method more robust and effective in ADMET classification settings, where tasks often exhibit varying degrees of label sparsity and heterogeneous supervision signals.

\section{Experiments and Analysis}
\subsection{Experimental Setup}
We evaluate our method on 13 classification tasks from TDC leaderboard, covering various pharmacological properties, including absorption, distribution, metabolism, and toxicity. We note that tasks related to excretion (the ``E" in ADMET) are primarily regression tasks, and thus are excluded from the scope of this study.

We adopt the leaderboard-style scaffold-based split strategy provided by TDC. Specifically, each dataset is divided into five folds, grouping molecules with similar core structures into the same subset. Each fold contains a \textit{train} and \textit{validation} subset, and a shared official \textit{test} set is provided separately for each task. To construct our multi-task dataset, we align the folds across all 13 tasks—merging fold-level splits of each task—and append the corresponding TDC-provided \textit{test} set to ensure consistent evaluation. In the latest version of Chemprop (v2), we use a predefined \texttt{split} column to indicate the subset each sample belongs to\footnote{See Chemprop documentation: \url{https://chemprop.readthedocs.io/en/latest/}}. To avoid data split conflicts that could lead to label leakage and incorrect supervision, our final multi-task data format retains separate label columns and split annotations for each task. SMILES strings that appear in multiple tasks are kept as independent rows, rather than being merged into a single entry.

All models are trained using the Chemprop framework with consistent hyperparameters across baselines and ablation variants.

\begin{table*}[t]
\renewcommand{\thetable}{2}
\centering
\begin{tabular}{cccccc}
\toprule
        \textbf{Tasks} & \textbf{Metric} & \textbf{SOTA Models} & \textbf{SOTA Score} & \textbf{QW-MTL (Ours)} & \textbf{Our Rank} \\
\midrule
        HIA         &  AUROC   &  MapLight + GNN   & 0.989$\pm$0.001  &     0.990$\pm$0.002      &      1 \\
  CYP2C9 Substrate  &  AUPRC   &     ZairaChem     & 0.441$\pm$0.033  & 0.446$\pm$0.023 &      1 \\
  CYP2D6 Substrate  &  AUPRC   &    ContextPred    & 0.736$\pm$0.024  &      0.725$\pm$0.025      &   2 \\
  CYP3A4 Substrate  &  AUROC   &    CNN (DeepPurpose)     & 0.662$\pm$0.008  & 0.660$\pm$0.016 &      2 \\
        DILI        &  AUROC   &     ZairaChem     & 0.925$\pm$0.005  & 0.932$\pm$0.014 &      1 \\
\bottomrule
\end{tabular}
\caption{Top-2 performing tasks where QW-MTL achieves leading performance on the TDC leaderboard (as of May 2025).}
\label{tab:main_results}
\end{table*}

\subsection{Compared with Other Methods}  
Table~\ref{tab:contrast} presents a comparison between our proposed QW-MTL framework and the strong baseline across 13 ADMET classification tasks. As evidenced by the results, QW-MTL consistently outperforms ADMET-AI on the majority of tasks, achieving notable gains on several challenging endpoints. Notably, the most significant gains are observed on low-resource tasks such as CYP2C9 Substrate, CYP2D6 Substrate, and DILI—each achieving a relative improvement of approximately 7\%. These tasks, typically constrained by limited training data, benefit substantially from QW-MTL's ability to leverage information from related high-resource tasks. These results underscore the strong transferability of our framework and its capacity to enhance sample efficiency by exploiting shared inductive biases across diverse ADMET endpoints.
\begin{table}[H]
\renewcommand{\thetable}{1}
\centering
\small
\begin{tabular}{ccccc}
\toprule
              \textbf{Task} & \textbf{Data Scale} & \textbf{Baseline} & \textbf{QW-MTL} & \textbf{Gain} \\
\midrule
Bio\_ma & 640 &0.675±0.045 & 0.706±0.017 &  3\% \\
               HIA & 578 &0.981±0.001 & 0.989±0.002 &  1\% \\
               Pgp &1211&0.895±0.013 & 0.913±0.013 &  2\% \\
               BBB &1975 &0.905±0.008 & 0.909±0.002 &  1\% \\
 CYP2C9-I &12092 &0.759±0.005 & 0.769±0.006 &  1\% \\
 CYP2D6-I &13130 &0.671±0.008 & 0.684±0.009 &  3\% \\
 CYP3A4-I  &12328  &0.870±0.003 & 0.875±0.003 &  1\% \\
  CYP2C9-S  & 666&0.415±0.019 & 0.445±0.023 &  7\% \\
  CYP2D6-S  & 664&0.675±0.034 &  0.722±0.020 &  7\% \\
  CYP3A4 S  & 667&0.621±0.026 & 0.645±0.016 &  6\% \\
              hERG &648 &0.833±0.008 & 0.842±0.013 &  2\% \\
              Ames & 7255&0.840±0.006 & 0.839±0.004 & 0\% \\
              DILI & 475&0.873±0.016 & 0.932±0.014 &  7\% \\
\bottomrule
\end{tabular}
\caption{
Performance comparison between the baseline and our proposed QW-MTL across 13 ADMET classification tasks.  
Task names are abbreviated for compact display: ``Bioavailability\_ma'' $\rightarrow$ ``Bio\_ma'', ``Inhibition'' $\rightarrow$ ``-I'', ``Substrate'' $\rightarrow$ ``-S'', etc.
\textbf{Baseline} refers to the single-task Chemprop model with RDKit descriptors (Chemprop-RDKit-STL).  
\textbf{QW-MTL} extends this baseline by incorporating multi-task learning, quantum descriptors, and a learnable task weighting mechanism.
}
\label{tab:contrast}
\end{table}

\begin{table*}[htbp]
\renewcommand{\thetable}{4}
\centering
\renewcommand{\arraystretch}{1.1}
\begin{tabular}{ccccc}
\toprule
\textbf{Tasks} & \textbf{Multi-RDKit} & \textbf{Multi-RDKit+QC} & \textbf{Multi-RDKit+Learnable-}$\bm{\beta}$ & \textbf{Multi-RDKit+QC+Learnable-}$\bm{\beta}$ \\
\midrule
Bioavailability\_ma & $0.705\pm 0.012$ & \bm{$0.709\pm 0.021$} & $0.692\pm 0.011$ & $0.695\pm 0.017$ \\
HIA & $0.989\pm 0.003$ & $0.989\pm 0.001$ & $0.988\pm 0.001$ & \bm{$0.990\pm 0.002$} \\
Pgp & $0.914\pm 0.010$ & \bm{$0.920\pm 0.008$} & $0.911\pm 0.013$ & \bm{$0.915\pm 0.013$} \\
BBB & $0.911\pm 0.003$ & $0.907\pm 0.007$ & $0.909\pm 0.009$ & $0.910\pm 0.002$ \\
CYP2C9 Inhibition & $0.769\pm 0.005$ & \bm{$0.770\pm 0.005$} & $0.767\pm 0.002$ & \bm{$0.770\pm 0.006$} \\
CYP2D6 Inhibition & $0.689\pm 0.009$ & \bm{$0.693\pm 0.005$} & $0.688\pm 0.008$ & \bm{$0.693\pm 0.009$} \\
CYP3A4 Inhibition & $0.871\pm 0.004$ & \bm{$0.873\pm 0.006$} & \bm{$0.872\pm 0.004$} & \bm{$0.877\pm 0.003$} \\
CYP2C9 Substrate & $0.456\pm 0.021$ & $0.434\pm 0.028$ & $0.449\pm 0.027$ & $0.446\pm 0.023$ \\
CYP2D6 Substrate & $0.721\pm 0.017$ & $0.721\pm 0.009$ & \bm{$0.724\pm 0.014$} & \bm{$0.725\pm 0.021$} \\
CYP3A4 Substrate & $0.649\pm 0.015$ & \bm{$0.651\pm 0.018$} & $0.637\pm 0.021$ & \bm{$0.660\pm 0.016$} \\
hERG & $0.843\pm 0.014$ & \bm{$0.855\pm 0.011$} & \bm{$0.849\pm 0.015$} & \bm{$0.847\pm 0.013$} \\
Ames & $0.835\pm 0.006$ & \bm{$0.839\pm 0.003$} & \bm{$0.836\pm 0.004$} & \bm{$0.836\pm 0.004$} \\
DILI & $0.928\pm 0.012$ & \bm{$0.930\pm 0.016$} & \bm{$0.939\pm 0.011$} & \bm{$0.938\pm 0.014$} \\
\bottomrule
\end{tabular}
\caption{Ablation results across different model variants. Bold numbers indicate performance improvements over the basic configuration (Multi-RDKit).}
\label{tab:ablation}
\end{table*}

Beyond outperforming our baseline, QW-MTL also demonstrates strong competitiveness on the TDC leaderboard. Table~\ref{tab:main_results} presents tasks on which our proposed QW-MTL framework achieves top performance on the TDC leaderboard (as of May 2025).  
For complete results across all 13 tasks, please refer to the \textbf{Appendix A}.  
Among the 13 classification tasks, QW-MTL achieves top-1 performance on 3 tasks and ranks top-2 on 5 tasks. Such performance, achieved on a highly competitive benchmark using a single compact model, underscores the practical value and effectiveness of QW-MTL as a general-purpose ADMET predictor. Moreover, this also highlights the benefits of multi-task learning when combining quantum-informed molecular descriptors with a learnable task weighting mechanism, and further validates the generalization ability and versatility of QW-MTL in molecular property prediction tasks characterized by significant task heterogeneity.

\subsection{Efficiency Comparison}
Despite introducing quantum chemical descriptors and a learnable multi-task weighting mechanism, QW-MTL maintains a comparable parameter count to the baseline Single-RDKit model (384k \textit{vs}. 378k).  
In addition to its strong predictive performance, QW-MTL significantly improves inference efficiency: on 10{,}000 molecules, it reduces inference time from 640.06 seconds to 60.88 seconds—a 10.5× speedup—under identical hardware and software conditions (detailed in the \textbf{Appendix B}).  
These results are summarized in Table~\ref{tab:efficiency}, demonstrating that QW-MTL offers a favorable trade-off between scalability, inference speed, and predictive accuracy without increasing model complexity.
\begin{table}[H]
\renewcommand{\thetable}{3}
\centering
\begin{tabular}{ccc}
\toprule
\textbf{Model} & \textbf{Single-RDKit} & \textbf{QW-MTL (Ours)} \\
\midrule
\textbf{Parameters} & 378,304 & 384,353 \\
\textbf{Inference Time (s)} & 640.06 & 60.88 \\
\bottomrule
\end{tabular}
\caption{Comparison of parameters and inference time.}
\label{tab:efficiency}
\end{table}
Furthermore, compared to other leading models in the ADMET prediction domain (see \textbf{Appendix C}), QW-MTL achieves competitive or superior performance with much lower computational demands.  
While state-of-the-art GNN-based or ensemble methods often rely on high-dimensional inputs (e.g., over 2,000 features) and large parameter sizes (typically exceeding 10M), QW-MTL achieves strong performance using only 508-dimensional input and 384k parameters.
\subsection{Ablation Study} 
\begin{figure}[H]
  \centering
  \includegraphics[width=0.98\linewidth]{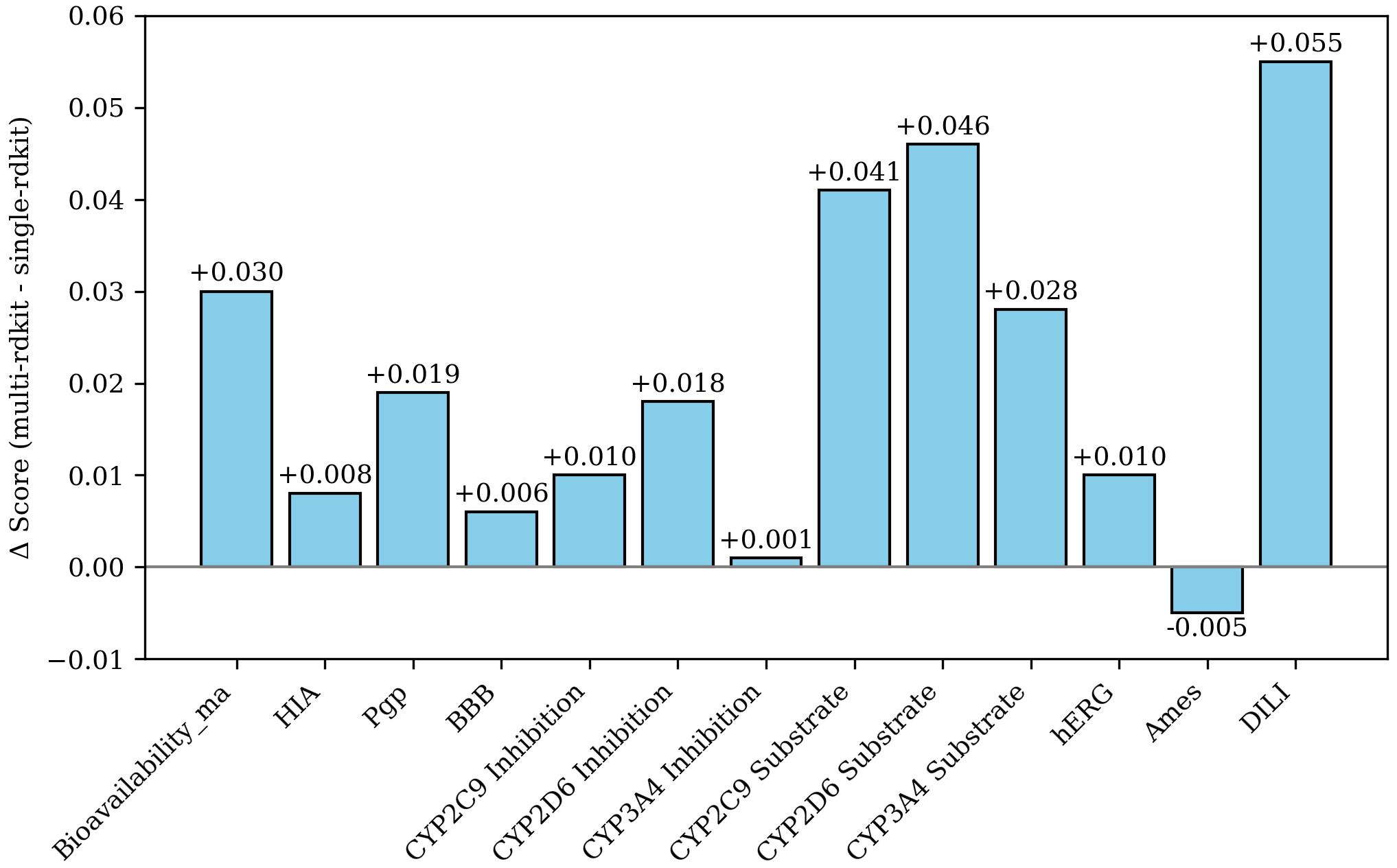}
  \caption{Performance gain of multi-task over single-task.}
  \label{fig:single-multi}
\end{figure}

We conduct a systematic ablation study to assess the contribution of each module in QW-MTL, including: (1) the strategy of using multitask framework, (2) the incorporation of quantum chemical descriptors (QC), (3) the design of the learnable task weighting mechanism $\beta$, and (4) the impact of full-task joint training compared to training on selected task subsets. To investigate the effectiveness of MTL strategy in isolation, we compare a single-task baseline (single-rdkit) with its multi-task counterpart (multi-rdkit), where both models are trained solely with RDKit descriptors. As shown in Figure~\ref{fig:single-multi}, we can see that the multi-rdkit model achieves consistently better performance across the majority of ADMET classification tasks, outperforming single-rdkit on 11 out of 13 endpoints. These results highlight the inherent benefit of multi-task training in leveraging cross-task correlations and shared molecular representations, providing a solid foundation for further architectural enhancements in subsequent modules.
To disentangle the contributions of quantum-informed molecular descriptors and learnable task weighting, we conduct an ablation study on all 13 ADMET classification tasks. As shown in Table~\ref{tab:ablation}, we compare four variants of the MTL framework: (1) a basic configuration that uses only RDKit features (\textbf{Multi-RDKit}), (2) the incorporation of quantum descriptors (+QC), (3) the introduction of learnable task-wise loss weights (+Learnable-$\beta$), and (4) the combination of both (+QC+Learnable-$\beta$), which corresponds to our \textbf{QW-MTL} framework.

Adding quantum descriptors leads to performance improvements on 9 out of 13 tasks relative to the baseline, with no significant degradation on the remaining tasks. In contrast, using learnable task weighting alone improves performance on 5 tasks but exhibits less consistent behavior overall. Notably, the full model improves performance on 10 out of 13 tasks and achieves the highest average score among all model variants, highlighting the complementary strengths of quantum descriptors and learnable task weighting.

These results suggest that quantum descriptors enhance the physical fidelity of learned embeddings, while task-specific weighting helps mitigate task imbalance. Their integration leads to more robust and consistent performance across diverse ADMET properties.
\begin{figure*}[!t]
  \renewcommand{\thefigure}{4}
  \centering
  \includegraphics[width=0.98\linewidth]{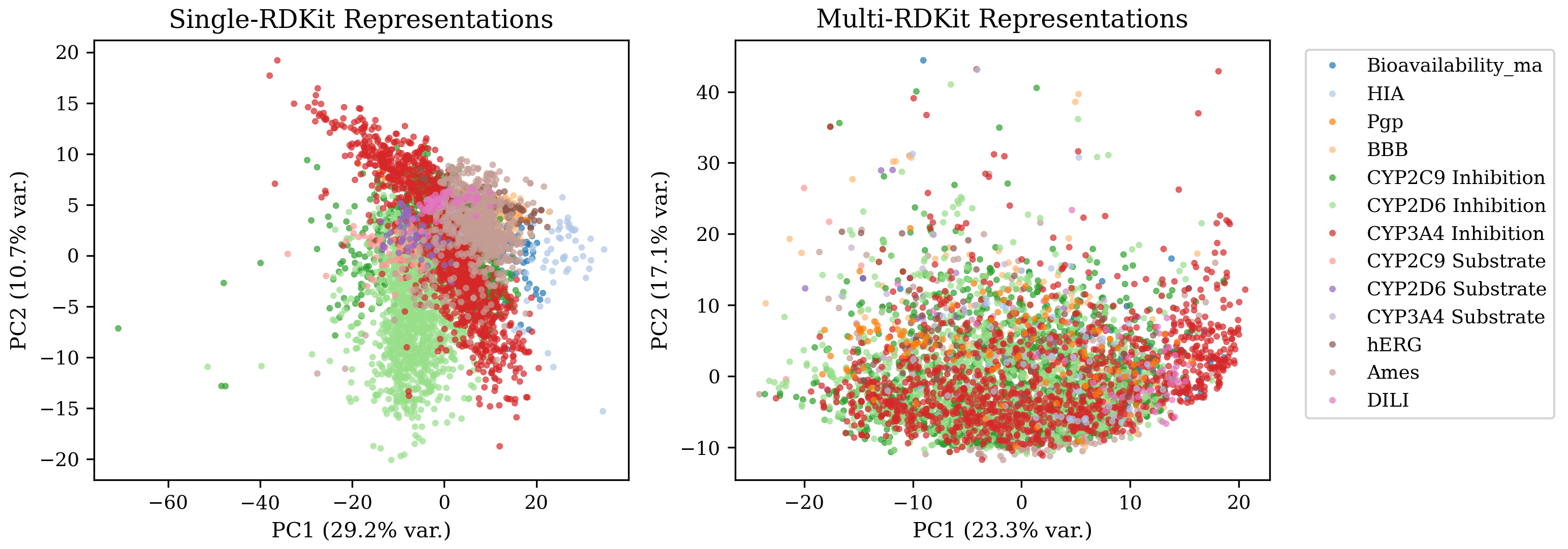}
  \caption{ Comparison of Single-RDKit and Multi-RDKit molecular representations using PCA.
}
  \label{fig:pca_single_vs_multi}
\end{figure*}
\subsection{Sample-Aware Task Weighting Analysis} 
In this section, we analyze the behavior of the learned exponents $\beta_t$—the task-specific versions of the global weighting parameter $\beta$—and investigate how they correlate with task-specific sample sizes and performance. We also evaluate the impact of our weighting strategy on overall multi-task optimization.

\begin{figure}[H]
    \centering
    \renewcommand{\thefigure}{3}
    \includegraphics[width=0.45\textwidth]{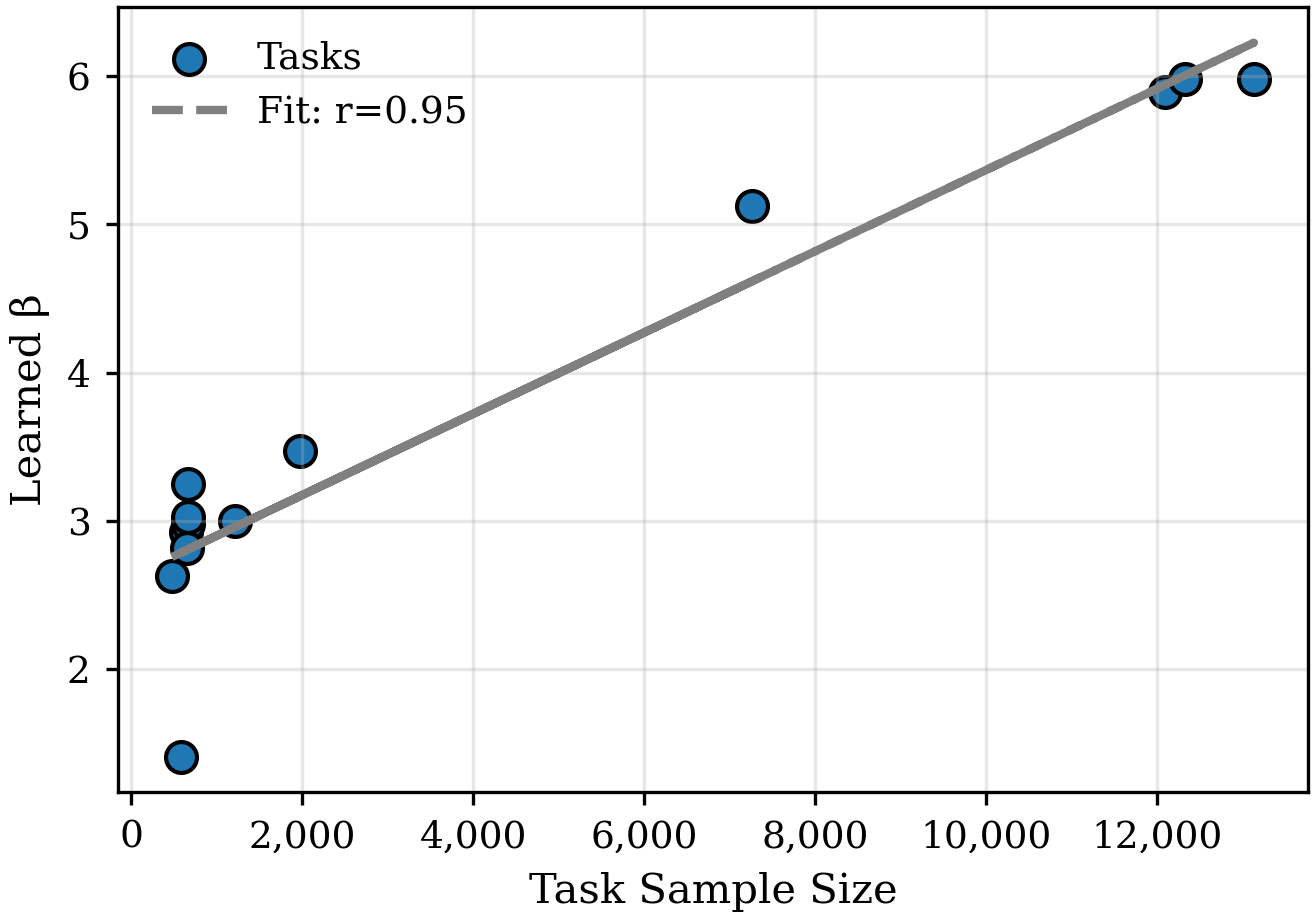}
    \caption{Correlation between task sample size and the learned exponent $\beta_t$. }
    \label{fig:beta_vs_scale}
\end{figure}

To assess whether the model learns to adapt $\beta_t$ according to the sample size of each task, we plot the relationship between task sample size and the corresponding learned $\beta_t$ values in Figure~\ref{fig:beta_vs_scale}. A strong positive correlation is observed, with a Pearson correlation coefficient of $r = 0.950$ ($p < 10^{-6}$), indicating that the model consistently adjusts the sharpness of the weighting function based on task scale.

Tasks with larger sample sizes tend to receive higher $\beta_t$ values, resulting in a steeper decay in $w_t = r_t^{\beta_t}$ and thus reducing their dominance in the overall loss. Conversely, tasks with smaller datasets are assigned lower $\beta_t$ values, effectively flattening their decay curves and ensuring their contributions remain non-negligible. This adaptive behavior aligns with the design objective of our weighting scheme: allowing task weights to reflect both relative scale and task importance.
\subsection{Visualization of Learned Representations}  
Since all single-task models are trained from the same initialization and share identical architectures,  
the resulting encoder representations are naturally aligned in the latent space, enabling consistent and meaningful comparisons across tasks. We adopt Principal Component Analysis (PCA) as the primary method for visualizing the learned molecular representations,  
as it preserves global structure and supports cross-model interpretation.  

Figure~\ref{fig:pca_single_vs_multi} presents a side-by-side PCA visualization comparing encoder representations obtained from single-task and multi-task models (Additional PCA for other experimental settings are included in \textbf{Appendix D}).  
The single-task models produce more dispersed and task-specific embeddings, with clear inter-task separation. In contrast, the multi-task model yields a more compact and overlapping distribution, suggesting the emergence of shared latent structures across tasks. Besides, we provide t-distributed Stochastic Neighbor Embedding (t-SNE) plots in the \textbf{Appendix E} to qualitatively examine local neighborhood structures and clustering behaviors. This entanglement reflects the model’s ability to discover common representational patterns that benefit multiple tasks simultaneously. Rather than being a drawback, the reduced separability indicates improved representational efficiency and stronger task transferability—hallmarks of effective multi-task learning. Such shared representations are particularly advantageous when tasks exhibit complementary characteristics or overlapping molecular mechanisms, as is often the case in ADMET prediction.
\section{Conclusion}
This work introduces \textbf{QW-MTL}, a unified multi-task learning framework for ADMET classification that integrates quantum-informed molecular descriptors with a learnable task weighting strategy. Built upon the Chemprop-RDKit backbone, QW-MTL effectively addresses task heterogeneity and data imbalance through physics-driven feature enhancement and dynamic optimization.

Experiments on 13 benchmark tasks from the TDC dataset show that QW-MTL consistently outperforms strong baselines, achieving robust performance while maintaining similar model complexity and faster inference time. Ablation studies further demonstrate the complementary benefits of quantum descriptors and sample-aware weighting, while latent space visualizations reveal more compact and transferable task representations.

Together, these findings establish QW-MTL as a \textbf{scalable and efficient} solution for multi-task ADMET prediction, well suited for large-scale molecular screening and property evaluation. Future directions include incorporating uncertainty-aware inference, pretraining on large molecular corpora, and designing task-specific architectures to better capture task diversity and interdependencies.
\clearpage
\bibliography{references}

\clearpage
\onecolumn
\begin{center}
\vspace*{2em}
{\Huge \textbf{Appendix}}\\[1ex]
\vspace*{2em}
\end{center}
\section{Appendix A}
\subsection{Additional Leaderboard Results}
Table~\ref{tab:leaderboard_all} provides the full leaderboard comparison across all 13 ADMET classification tasks from the TDC benchmark, based on leaderboard results as of May 2025.
\begin{table}[H]
\centering
\renewcommand{\thetable}{1}
\begin{tabular}{cccccc}
\toprule
\textbf{Task} & \textbf{Metric} & \textbf{SOTA Model} & \textbf{SOTA Score} & \textbf{QW-MTL (Ours)} & \textbf{Our Rank} \\
\midrule
Bioavailability\_ma & AUROC & SimGCN & 0.748±0.033 & 0.695±0.017 & 6 \\
HIA & AUROC & MapLight + GNN & 0.989±0.001 & 0.990±0.002 & 1 \\
Pgp & AUROC & MapLight + GNN & 0.938±0.002 & 0.915±0.013 & 8 \\
BBB & AUROC & MapLight & 0.916±0.001 & 0.910±0.002 & 5 \\
CYP2C9 Inhibition & AUPRC & MapLight + GNN & 0.859±0.001 & 0.770±0.006 & 8 \\
CYP2D6 Inhibition & AUPRC & MapLight + GNN & 0.790±0.001 & 0.693±0.009 & 5 \\
CYP3A4 Inhibition & AUPRC & MapLight + GNN & 0.916±0.000 & 0.877±0.003 & 5 \\
CYP2C9 Substrate & AUPRC & ZairaChem & 0.441±0.033 & 0.446±0.023 & 1 \\
CYP2D6 Substrate & AUPRC & ContextPred & 0.736±0.024 & 0.725±0.021 & 2 \\
CYP3A4 Substrate & AUROC & CNN (DeepPurpose) & 0.662±0.008 & 0.660±0.016 & 2 \\
hERG & AUROC & MapLight + GNN & 0.880±0.002 & 0.847±0.013 & 4 \\
Ames & AUROC & ZairaChem & 0.918±0.001 & 0.912±0.002 & 3 \\
DILI & AUROC & MolBERT & 0.878±0.007 & 0.890±0.007 & 1 \\
\bottomrule
\end{tabular}
\caption{Performance Comparison on TDC Leaderboard Tasks.}
\label{tab:leaderboard_all}
\end{table}
\section{Appendix B}
\subsection{Inference Efficiency Evaluation}
\textbf{Molecule Selection.}
All inference efficiency measurements were conducted on a shared test set consisting of 10,000 molecules sampled from the union of TDC test splits across all 13 tasks.\\
\textbf{Hardware Setup.}
Inference time was measured on a single MIG 40G partition of an NVIDIA A800-SXM4-80GB GPU within an AI-X86 NVIDIA cluster node equipped with 4 CPU cores. All models were evaluated using PyTorch 2.7.0 and CUDA 12.9, under a consistent software environment. \\
\textbf{Software Setup.}
All evaluations were performed using Chemprop’s default inference settings to ensure consistency across model variants~\cite{43}. 
\textbf{Results.}
To ensure robustness, we conducted inference benchmarking three times on independently scheduled but identically configured GPUs. Compared to the baseline Single-RDKit model, QW-MTL achieved speedup factors of 10.5×, 14×, and 12× across the three runs. The main text reports the most conservative result (10.5×), while all measurements are summarized in Table~\ref{tab:inference_time}. These results demonstrate that QW-MTL substantially improves inference efficiency while maintaining strong predictive performance, and remains lightweight even with added descriptor complexity.
\begin{table}[H]
\centering
\renewcommand{\thetable}{2}
\begin{tabular}{cccc}
\toprule
\textbf{Run} & \textbf{Single-RDKit (Baseline)} & \textbf{QW-MTL (Ours)} & \textbf{Speedup ($\times$)} \\
\midrule
Run 1 & 640.06s & 60.88s & 10.51 \\
Run 2 & 702.31s & 49.15s & 14.29 \\
Run 3 & 669.85s & 52.37s & 12.79 \\
\bottomrule
\end{tabular}
\caption{Inference Time and Speedup Comparison Across Runs.}
\label{tab:inference_time}
\end{table}
\section{Appendix C}
The 508-dimensional input of QW-MTL mentioned in the main text represents the full molecular representation (208 external descriptors + 300 learned embedding).
\begin{table}[htbp]
\centering
\label{tab:model_dim_comparison}
\begin{tabular}{ccc}
\toprule
\textbf{Model} & \textbf{\#Params} & \textbf{Dim for Prediction} \\
\midrule
QW-MTL (Ours)                        & 384K     & 508   \\
MapLight+GNN                         & --\footnotemark[1] & 2863  \\
AttrMasking                          & 2.07M    & --\footnotemark[1] \\
SimGCN                               & 1.10M    & 200   \\
ContextPred                          & 2.07M    & 300   \\
RDKit2D + MLP (DeepPurpose)          & 633K     & 200   \\
Morgan + MLP (DeepPurpose)           & 148K     & 1024  \\
\bottomrule
\end{tabular}
\caption{Comparison of parameter counts and representation dimensions used for prediction. “–” indicates information not reported in the original source.}
\footnotetext[1]{Information not reported in original source.}
\end{table}
\section{Appendix D}
\subsection{PCA Visualizations for All Model Settings}
To further investigate the latent representations learned under different model configurations, we provide PCA visualizations for all five settings: Single-RDKit, Multi-RDKit, Multi-RDKit+QC, Multi-RDKit+Learnable-$\beta$, and Multi-RDKit+QC+Learnable-$\beta$.  
Each model was trained on the same set of molecules across 13 ADMET classification tasks, and PCA was applied separately to the encoder outputs obtained on a shared validation set. Specifically, we select molecules from the \texttt{fold-1} validation split of each dataset as a representative evaluation set to ensure consistency across tasks and settings.

\begin{figure}[H]
    \centering
    \includegraphics[width=\textwidth]{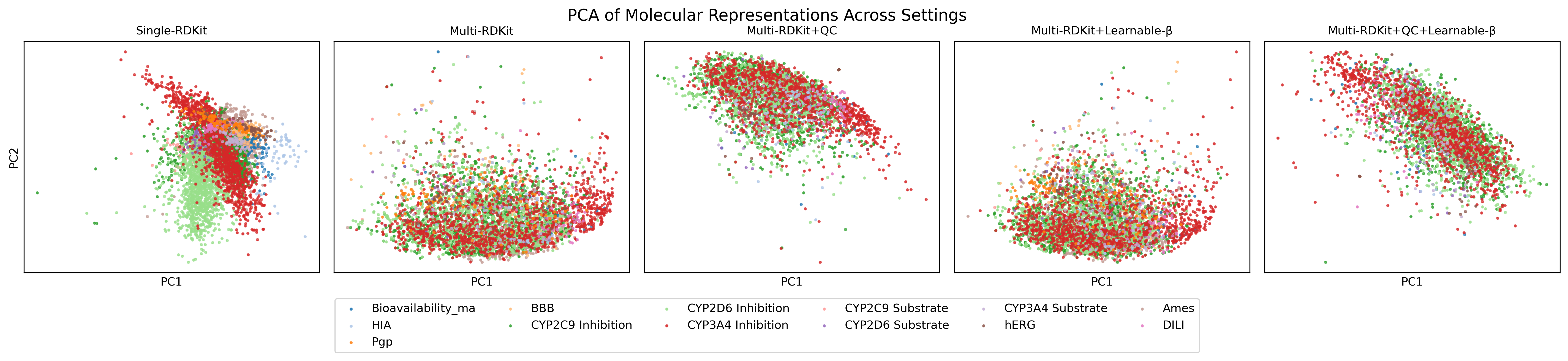}
    \caption{PCA visualizations of molecular representations under different model settings. Each subfigure corresponds to one model variant trained on the same molecule set across 13 ADMET classification tasks. Representations are extracted from the fold-1 validation subset.}
    \label{fig:pca_all_settings}
\end{figure}

As shown in Figure~\ref{fig:pca_all_settings}, the visualizations reveal how different modeling choices influence the geometry and compactness of the learned molecular embeddings.
\section{Appendix E}
\subsection{t-SNE Visualizations of Molecular Representations}
\begin{figure}[H]
    \centering
    \includegraphics[width=\textwidth]{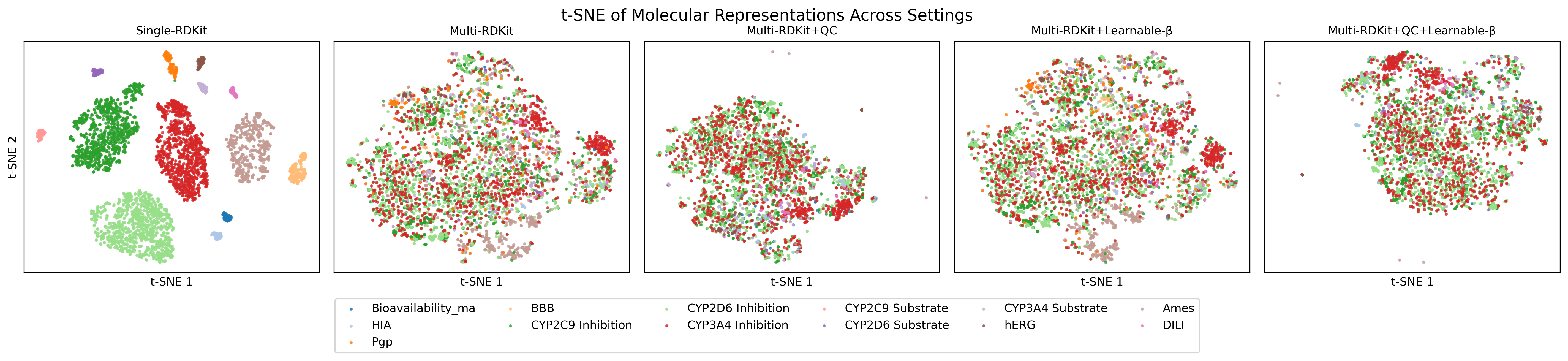}
    \caption{t-SNE visualizations of molecular representations under different model settings. Each subfigure corresponds to one model variant trained on the same molecule set across 13 ADMET classification tasks. Representations are extracted from the fold-1 validation subset.}
    \label{fig:tsne_all_settings}
\end{figure}

While PCA captures global geometry, t-SNE focuses on preserving local neighborhoods, allowing us to qualitatively assess clustering behaviors and local structure in the latent space. As shown in Figure~\ref{fig:tsne_all_settings}, Across different model settings, we observe varying degrees of task-wise clustering and overlap.
\section{Appendix F}
Table~\ref{tab:fig2_data} presents the numerical values used to generate Figure~2 in the main text.
\begin{table}[H]
\centering
\renewcommand{\thetable}{4}
\begin{tabular}{ccc}
\toprule
\textbf{Task} & \textbf{Single-RDKit} & \textbf{Multi-RDKit} \\
\midrule
Bioavailability\_ma & 0.621 & 0.649 \\
HIA                 & 0.675 & 0.705 \\
Pgp                 & 0.833 & 0.843 \\
BBB                 & 0.895 & 0.914 \\
CYP2C9 Inhibition   & 0.981 & 0.989 \\
CYP2D6 Inhibition   & 0.905 & 0.911 \\
CYP3A4 Inhibition   & 0.840 & 0.835 \\
CYP2C9 Substrate    & 0.873 & 0.928 \\
CYP2D6 Substrate    & 0.671 & 0.689 \\
CYP3A4 Substrate    & 0.415 & 0.456 \\
hERG                & 0.759 & 0.769 \\
Ames                & 0.675 & 0.721 \\
DILI                & 0.870 & 0.871 \\
\bottomrule
\end{tabular}
\caption{Performance comparison between Single-RDKit and Multi-RDKit on 13 ADMET classification tasks.}
\label{tab:fig2_data}
\end{table}

Table~\ref{tab:beta_values} provides the raw data used to generate Figure 3 in the main text.
\begin{table}[H]
\centering
\renewcommand{\thetable}{5}
\begin{tabular}{ccc}
\toprule
\textbf{Task} & \textbf{Data Scale} & \boldmath$\boldsymbol{\beta}$ \\
\midrule
Bioavailability\_ma & 640 & 5.123 \\
HIA                 & 578 & 3.469 \\
Pgp                 & 1211 & 2.928 \\
BBB                 & 1975 & 2.981 \\
CYP2C9 Inhibition   & 12092 & 5.892 \\
CYP2D6 Inhibition   & 13130 & 3.027 \\
CYP3A4 Inhibition   & 12328 & 2.907 \\
CYP2C9 Substrate    & 666 & 6.000 \\
CYP2D6 Substrate    & 664 & 5.978 \\
CYP3A4 Substrate    & 667 & 2.630 \\
hERG                & 648 & 2.820 \\
Ames                & 7255 & 1.412 \\
DILI                & 475 & 3.000 \\
\bottomrule
\end{tabular}
\caption{Learned $\beta$ values and data scale for each task.}
\label{tab:beta_values}
\end{table}
\end{document}